\documentclass[11pt,a4paper]{article}
\usepackage[hyperref]{emnlp2020}
\usepackage{times}
\usepackage{latexsym}
\usepackage{graphicx}
\usepackage{amsmath}
\usepackage{amssymb}
\usepackage{multirow}
\usepackage{pgfplots}
\usepackage{booktabs}
\usepackage{subfigure}
\usepackage{xcolor}
\usepackage{comment}
\usepackage{bm}
\usepackage{url}
\pgfplotsset{width=0.9\columnwidth,compat=1.9}

\usepackage{microtype}

\definecolor{mygreen}{rgb}{0.372822,0.630254,0.217202}
\definecolor{myblue}{rgb}{0.208317,0.357233,0.717412}
\definecolor{myorange}{rgb}{1,0.5,0}
\aclfinalcopy

\title{Language Generation with Multi-Hop Reasoning on Commonsense Knowledge Graph}

\author{Haozhe Ji$^1$, Pei Ke$^1$, Shaohan Huang$^2$, Furu Wei$^2$, Xiaoyan Zhu$^1$, Minlie Huang$^1$\thanks{\quad Corresponding author}  \\
$^1$Department of Computer Science and Technology,
Institute for Artificial Intelligence, \\
State Key Lab of Intelligent Technology and Systems, \\ 
Beijing National Research Center for Information Science and Technology, \\ 
Tsinghua University, Beijing 100084, China \\
$^2$Microsoft Research \\
  {\tt\small \{jhz20,kp17\}@mails.tsinghua.edu.cn, \{shaohanh, fuwei\}@microsoft.com, } \\
  {\tt\small \{zxy-dcs,aihuang\}@tsinghua.edu.cn } \\}

\date{}

\begin{document}
\maketitle
\begin{abstract}
Despite the success of generative pre-trained language models on a series of text generation tasks, they still suffer in cases where reasoning over underlying commonsense knowledge is required during generation.
Existing approaches that integrate commonsense knowledge into generative pre-trained language models simply transfer relational knowledge by post-training on individual knowledge triples while ignoring rich connections within the knowledge graph. We argue that exploiting both the structural and semantic information of the knowledge graph facilitates commonsense-aware text generation. In this paper, we propose \textit{Generation with Multi-Hop Reasoning Flow} (GRF) that enables pre-trained models with dynamic multi-hop reasoning on multi-relational paths extracted from the external commonsense knowledge graph. We empirically show that our model outperforms existing baselines on three text generation tasks that require reasoning over commonsense knowledge. We also demonstrate the effectiveness of the dynamic multi-hop reasoning module with reasoning paths inferred by the model that provide rationale to the generation.\footnote{The source code is available at \url{https://github.com/cdjhz/multigen}.}
\end{abstract}

\section{Introduction}
\begin{figure}
    \centering
    \includegraphics[width=1.0\columnwidth]{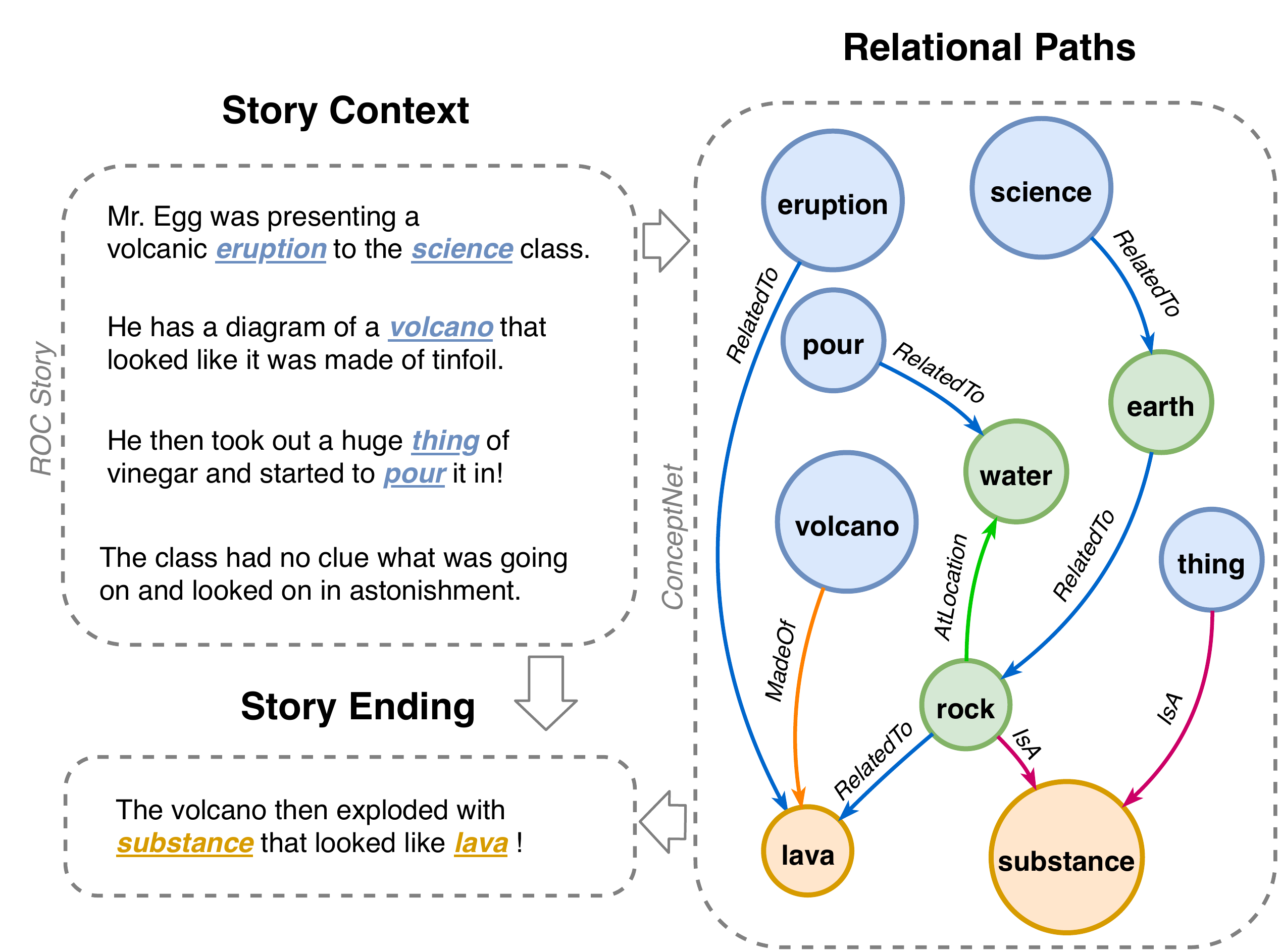}
    \caption{An example of using structural relational knowledge as commonsense grounding in story ending generation. \textcolor{myblue}{Blue nodes} correspond to the concepts in the context, \textcolor{myorange}{orange nodes} correspond to those in the
    story ending and \textcolor{mygreen}{gree nodes} are intermediate concepts that connect the evidence chain.}
    \label{fig:example}
\end{figure}

Despite the recent success of pre-trained language models such as GPT-2~\citep{Radford2019LanguageMA} on various language generation tasks, these models are still struggling on generation tasks that require reasoning over commonsense knowledge that is not \textit{explicitly} stated in the context. 
For example, Figure \ref{fig:example} illustrates an example in the story ending generation task, where external commonsense knowledge in the form of relational paths 
can guide the generation of the key concepts ``substance'' and ``lava'' in the story ending by providing background knowledge such as $(\texttt{volcano}, \texttt{MadeOf}, \texttt{lava})$ besides the story context. Although pre-trained models have been demonstrated to 
possess commonsense 
reasoning ability~\citep{Trinh2018ASM} by \textit{implicitly} learning some relational patterns from large-scale corpora, they do not fully utilize the commonsense knowledge bases that provide more explicit knowledge grounding.


To address this defect, incorporating external commonsense knowledge to enhance models' reasoning ability has been widely explored~\citep{lin-etal-2019-kagnet,Ye2019AlignMA,Lv2020GraphBasedRO}.
In language generation, previous work~\citep{Bhagavatula2019AbductiveCR, guan2020knowledge} transfers commonsense knowledge into pre-trained language models by utilizing triple information in commonsense knowledge bases such as ConceptNet~\citep{Speer2012RepresentingGR} and ATOMIC~\citep{Sap2019ATOMICAA}. 

However, this approach has two drawbacks. 
First, recovering knowledge triples at the post-training stage~\citep{guan2020knowledge} hardly enables the model to utilize the encoded knowledge in fine-tuning generation tasks which requires reasoning over underlying commonsense knowledge.
Second, it ignores the abundant structural relational relevance of the concepts in the knowledge graph~\citep{guan2020knowledge,Bhagavatula2019AbductiveCR} that may provide multiple plausible evidence for complex reasoning. 
Thus a richer and more explicit way of utilizing external commonsense knowledge is to exploit both \textit{structural} and \textit{semantic} information of the knowledge graph and reason over multi-hop relational paths where multiple connected triples provide chains of evidence for grounded text generation.

\begin{figure*}[t]
    \centering
    \includegraphics[width=1.7\columnwidth]{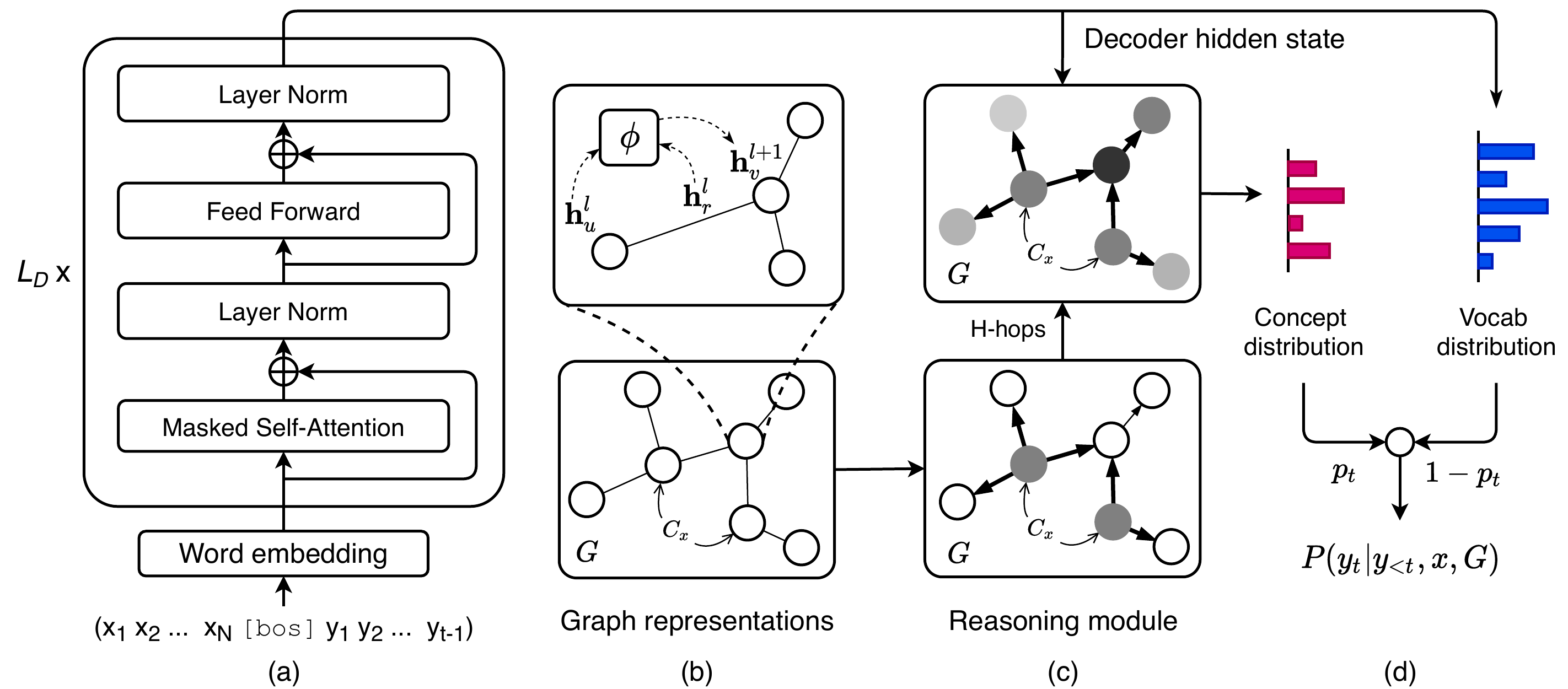}
    \caption{Model architecture. (a) Context modeling with pre-trained transformer (\S{\ref{sec:text}}). (b) The model encodes the multi-relational graph with non-parametric operation $\phi(\cdot)$ to combine relations and concepts (\S{\ref{sec:static-graph}}). 
    (c) The multi-hop reasoning module aggregates evidence from source concepts $C_{\bm{x}}$ along structural paths to all nodes where shade indicates the node score (\S{\ref{sec:dmrf}}). (d) 
   The final generation distribution with gate control (\S{\ref{sec:gate}}).}
    \label{fig:model}
\end{figure*}

In this paper, we propose \textit{Generation with Multi-Hop Reasoning Flow} (GRF), a generation model that performs multi-hop reasoning on the external knowledge graph for knowledge-enriched language generation.
The model operates on the sub-graph extended from the concepts in the input text as commonsense knowledge grounding. It first encodes the multi-relational graph with compositional operation to obtain graph-aware representations for the concepts and the relations (\S{\ref{sec:static-graph}}). 
Then, the multi-hop reasoning module performs dynamic reasoning via aggregating triple evidence along multiple relational paths 
to generate the salient concept under the context (\S{\ref{sec:dmrf}}). 
Finally, the generation distribution combines the probability of copying concepts from the knowledge graph and that of choosing a word from the standard vocabulary with a gate control (\S{\ref{sec:gate}}). The overall model architecture is shown in Figure \ref{fig:model}. 
We conduct experiments on three commonsense-aware text generation tasks
including story ending generation~\citep{Mostafazadeh2016ACA}, abductive natural language generation~\citep{Bhagavatula2019AbductiveCR}, and explanation generation for sense making~\citep{Wang2019DoesIM}. 
Results show that our model outperforms strong baselines on these tasks, thereby demonstrating the benefit of multi-hop commonsense reasoning in language generation.



Our contributions can be summarized as follows: 1) We propose GRF, a novel generation model that utilizes external structural commonsense knowledge to facilitate explicit commonsense reasoning in text generation. 2) We propose the dynamic multi-hop reasoning module that 
aggregates evidence along relational paths for grounded generation of some critical concepts.
3) We conduct extensive experiments including automatic and human evaluation on three commonsense-aware text generation tasks and show that our model outperforms various selective baselines. We also visualize reasoning paths inferred by the model to demonstrate the effectiveness of the multi-hop reasoning module. 


\section{Related Work}
\subsection{Commonsense-Aware Neural Text Generation}
Incorporating commonsense knowledge is essential for text generation to 
augment the limited textual information. In dialogue generation, \citet{Zhou2018CommonsenseKA} enriched the context representations of the post with neighbouring concepts on ConceptNet using graph attention. In story ending generation, \citet{Guan2019StoryEG} proposed incremental encoding with multi-source attention to incorporate one-hop knowledge graph for concepts in the story context.
In topic-to-essay generation, \citet{yang-etal-2019-enhancing-topic} augmented the generator with a concept memory that updated dynamically with gate mechanism.
Recently, some work also attempted to integrate external commonsense knowledge into generative pre-trained language models such as GPT-2~\citep{Radford2019LanguageMA}. \citet{guan2020knowledge} conducted post-training on sythetic data constructed from commonsense knowledge bases by translating triplets into natural language texts using templates. \citet{Bhagavatula2019AbductiveCR} transferred embeddings of COMeT~\citep{Bosselut2019COMETCT}, a GPT-2 model fine-tuned to 
generate the tail entity of a triple in commonsense knowledge graph, into another GPT-2 model for text generation. In comparison, our model utilizes both structural and semantic information of the commonsense knowledge graph during generation and does not suffers from the catastrophic forgetting problem~\citep{Kirkpatrick2016OvercomingCF} caused by implicit knowledge transferring.

\subsection{Multi-Hop Reasoning on Graph Structure}
Performing explicit multi-hop reasoning on graph structure has been demonstrated to be an effective approach for query answering over incomplete knowledge graphs~\citep{Das2018GoFA, Chen2018VariationalKG, Lin2018MultiHopKG}, multi-hop question answering~\citep{Bauer2018CommonsenseFG,Cao2018QuestionAB,Xiao2019DynamicallyFG} and dialogue generation~\citep{Tuan2019DyKgChatBD,Moon2019OpenDialKGEC, Liu2019KnowledgeAC}. Particularly, reasoning on knowledge graphs to answer relational query typically adopts REINFORCE to learn concrete policies to search for entities or relations. In multi-hop question answering tasks, the reasoning process is augmented with entity graph~\citep{Cao2018QuestionAB,Xiao2019DynamicallyFG} or concept paths~\citep{Bauer2018CommonsenseFG} to enhance semantic connections among document segments. 
In dialogue generation, \citet{Tuan2019DyKgChatBD} modeled multiple hops on relationship graphs with a Markov transition matrix. \citet{Liu2019KnowledgeAC} proposed a two-stage architecture that selected information from a knowledge graph for further generating the response. Compared with these generation models that operate on knowledge graphs within a specific domain, our focus is to utilize general commonsense knowledge to supply evidence for text generation.

\section{Methodology}

\subsection{Problem Formulation}

In this paper, we focus on text generation tasks where reasoning over external commonsense knowledge is required. Without loss of generality, the input source is a text sequence $\bm{x}=(x_1,x_2,\cdots, x_N)$ which may consist of several sentences. The output target is another text sequence $\bm{y}=(y_1,y_2,\cdots,y_M)$. 
To facilitate the reasoning process, we resort to an external commonsense knowledge graph $\mathcal{G}=(\mathcal{V}, \mathcal{E})$ where $\mathcal{V}$ denotes the concept set and $\mathcal{E}$ denotes the relations connecting these concepts. Since direct reasoning on the complete graph is intractable, we extract a sub-graph $G=(V,E)$ given the input text where $V\subset\mathcal{V}$ and $E\subset\mathcal{E}$. 
The sub-graph consists of inter-connected $H$-hop paths starting from the source concepts $C_{\bm{x}}$ extracted from the input text. We only consider concepts with $1$-gram surface texts. The task is then formulated as generating the best hypothesis $\hat{\bm{y}}$ which maximizes the following conditional probability:
\begin{equation}
    \hat{\bm{y}}=\text{argmax}_{\bm{y}}P(\bm{y}|\bm{x}, G).
\end{equation}

We leave the detailed sub-graph extraction process in \S{\ref{sec:subgraph}} and describe our proposed model in the next section.

\subsection{Generation with Multi-Hop Reasoning Flow}


\subsubsection{Static Multi-Relational Graph Encoding}\label{sec:static-graph}
Graph Neural Network (GNN) frameworks, such as graph convolution network (GCN)~\citep{Kipf2017SemiSupervisedCW} and graph attention network (GAT)~\citep{Velickovic2018GraphAN}, have been shown effective at encoding graph-structured data by aggregating node information from local neighbours. 
To model the relational information in the knowledge graph, R-GCN~\citep{Schlichtkrull2018ModelingRD} generalizes GCN with relation-specific weight matrices but is reported to be over-parameterized~\citep{Marcheggiani2017EncodingSW, Schlichtkrull2018ModelingRD}. We follow \citet{Vashishth2019CompositionbasedMG} and use a non-parametric compositional operation $\phi(\cdot)$ to combine the node embedding and the relation embedding. Specifically, given the input graph $G=(V, E)$ and a GCN with $L_G$ layer, for each node $v\in V$ we update the node embedding at the $l+1$-th layer by aggregating information from its local neighbours $\mathcal{N}(v)$ which consist of pairs of node $u$ and the connected relation $r$.
\begin{align}
   \bm{o}_v^l &= \frac{1}{|\mathcal{N}(v)|} \sum_{(u, r)\in\mathcal{N}(v)}\mathbf{W}_N^l\phi(\bm{h}_u^l, \bm{h}_r^l), \\
    \bm{h}_v^{l+1}&=\text{ReLU}\Big(\bm{o}_v^l
    + \mathbf{W}_S^l \bm{h}_v^l \Big),
\end{align}
where $\bm{h}_v^0$ is initialized by looking up the word embedding and $\bm{h}_r^0$ by the relation-type embedding.
$\mathbf{W}_N^l$ and $\mathbf{W}_S^l$ are two learnable weight matrices specific to the $l$-th layer. 
We define the compositional operation as $\phi(\bm{h}_u, \bm{h}_r)=\bm{h}_u - \bm{h}_r$ 
inspired by the TransE model~\citep{Bordes2013TranslatingEF}. 

The relation embedding is also updated via another linear transformation.
\begin{equation}
    \bm{h}_r^{l+1}=\mathbf{W}_{R}^l \bm{h}_r^l.
\end{equation}

Finally, we obtain node embeddings $\bm{h}^{L_G}_v$ and relation embeddings $\bm{h}^{L_G}_r$ that encode the static graph context for dynamic reasoning during decoding. 

\subsubsection{Context Modeling with Pre-Trained Transformer}\label{sec:text}
We adopt the GPT-2 model~\citep{Radford2019LanguageMA}, a pre-trained multi-layer transformer decoder to model the contextual dependency of the text sequence. The input to the model is the concatenation of the source and the target sequence: $\bm{s} = (x_1,\cdots,x_N,[\texttt{bos}],y_1,\cdots,y_M)$.
\begin{align}
    \bm{h}_t^0 &= \bm{e}_t + \bm{p}_t, \\
    \bm{h}_t^l &= \texttt{T\_block}(\mathbf{H}_{\le t}^{l-1}), l\in[1,L_D] \\
    \label{equ:lm}
    P(s_t|\bm{s}_{<t}) &= \text{softmax}(\mathbf{W}_{LM}\bm{h}_t^{L_D} + \bm{b})
\end{align}
where $\bm{e}_t$ and $\bm{p}_t$ are the token embedding vector and the positional embedding vector.
$\texttt{T\_block}$ is the transformer block with masked self-attention. 
The final hidden state at the $t$-th time step $\bm{h}_t^{L_D}$ which encodes the context information is used as the input to the multi-hop reasoning module.


\subsubsection{Dynamic Multi-Hop Reasoning Flow}\label{sec:dmrf}

To perform explicit reasoning on the graph structure during generation, we devise a dynamic reasoning module that utilizes both structural patterns of the knowledge graph and contextual information to propagate evidence along relational paths at each decoding step. 

Specifically, the module broadcasts information on $G$ by updating the score of outer nodes with their visited neighbours for multiple hops until all the nodes on $G$ are visited. Initially, nodes correspond to the concepts in $C_{\bm{x}}$ are given a score of $1$ while other unvisited nodes are assigned with $0$.



For the unvisited node $v\in V$, its node score $ns(v)$ is computed by aggregating evidence from $\mathcal{N}_{in}(v)$ which denotes the set of visited node $u$ and its edge $r$ that directly connects $v$.
\begin{equation}
    ns(v) = \underset{(u, r)\in \mathcal{N}_{in}(v)}{f}\Big(\gamma \cdot ns(u) + R(u,r,v) \Big),
\end{equation}
where $\gamma$ is a discount factor that controls the intensity of the information flow from the previous hops. $f(\cdot)$ is the aggregator that assembles scores from connected nodes. We consider two forms of aggregators: $\texttt{max}(\cdot)$ and $\texttt{mean}(\cdot)$. We use $\texttt{max}(\cdot)$ for the main results and present the results with $\texttt{mean}(\cdot)$ in the ablation study. 

$R(u,r,v)$ is the triple relevance that reflects the relevancy of the evidence given by the triplet $(u, r, v)$ under the current context. We compute the triple relevance as follows:
\begin{align}
    R(u,r,v)&=\sigma(\bm{h}_{u,r,v}^\mathrm{T}\mathbf{W}_{sim} \bm{h}_t^{L_D}),\\
    \bm{h}_{u,r,v} &= [\bm{h}_u^{L_G}; \bm{h}_r^{L_G}; \bm{h}_v^{L_G}].
\end{align}

After $H$ hops, the final distribution over the nodes is obtained by a normalization.
\begin{equation}\label{equ:node}
    P(c_t|\bm{s}_{<t}, G)=\text{softmax}_{v\in V}(ns(v)),
\end{equation}
where $c_t$ is the concept of the selected node at the $t$-th time step. 

Intuitively, the reasoning module learns to dynamically distribute along the paths by considering the triple evidence according to the current decoder state.

\subsubsection{Generation Distribution with Gate Control}\label{sec:gate}
The final generation distribution combines the distribution over the concepts (Eq. \ref{equ:node}) and the distribution over the standard vocabulary (Eq. \ref{equ:lm}). We use a soft gate probability $g_t$ which denotes whether to copy a concept in the generation to control the weight of the two distributions similar to the copy mechanism~\citep{Gu2016IncorporatingCM,See2017GetTT}.
\begin{equation}
    g_t = \sigma\Big(\mathbf{W}_{gate}\bm{h}_t^{L_D}\Big).
\end{equation}
The final output distribution is the linear combination of the two distributions weighted by $g_t$ and $1-g_t$ respectively. 
\begin{align}
    P(y_t|\bm{y}_{<t},\bm{x}, G) &= g_{t+N} \cdot P(c_{t+N}|\bm{s}_{<t+N}, G) \nonumber \\ 
    &+ (1-g_{t+N}) \cdot P(s_{t+N}|\bm{s}_{<t+N}),
\end{align}
where $N$ is the length of the input text sequence.

\subsection{Training and Inference}
To train the proposed model, we minimize the negative log-likelihood of generating the ground truth target sequence $\bm{y}^{\text{gold}}=(y_1,y_2\cdots, y_M,[\texttt{eos}])$.
\begin{equation}
    \mathcal{L}_{gen} = \sum_{t=1}^{M+1} -\log P(y^{\text{gold}}_t|\bm{y}^{\text{gold}}_{<t},\bm{x}, G).
\end{equation}

We add an auxiliary gate loss $\mathcal{L}_{gate}$ to supervise the probability of selecting a concept or a generic word. We additionally introduce a weak supervision $\mathcal{L}_{weak}$ to induce the predicted triple relevances to match the heuristic labels of edges obtained by breadth-first search from the source concepts to the concepts in $\bm{y}^{\text{gold}}$ on the graph. 
Both loss functions take the form of binary cross-entropy. We observe that both loss terms encourage the model to learn multi-hop reasoning on the graph more effectively.


The final loss to be optimized is the linear combination $\mathcal{L}_{gen} + \alpha \mathcal{L}_{gate} + \beta \mathcal{L}_{weak}$.

During the inference stage, the input to the model is $(x_1,\cdots,x_N,[\texttt{bos}])$. The model generates a token at a time and concatenates it to the input sequence to generate the next token. The generation process terminates when the special ending symbol $[\texttt{eos}]$ is generated.

\section{Experiments}

\subsection{Datasets and Metrics}

The statistics of the datasets are shown in Table \ref{tab:stats}.

\noindent\textbf{Story Ending Generation} (SEG) is to generate a reasonable ending given a four-sentence story context. The stories come from ROCStories corpus~\citep{Mostafazadeh2016ACA}. We use the same data split as \citet{Guan2019StoryEG}.

\noindent\textbf{Abductive NLG} ($\alpha$NLG) is to generate an explanatory hypothesis given two observations: $O_1$ as the cause and $O_2$ as the consequence. We use the official data split\footnote{\url{https://github.com/allenai/abductive-commonsense-reasoning}} from \citet{Bhagavatula2019AbductiveCR}.

\noindent\textbf{Explanation Generation} (EG) is to generate an explanation given a counter-factual statement for sense-making~\citep{Wang2019DoesIM}. We randomly split $85\%$ of the data as the training set, $10\%$ as the test set, and the latter as the development set.

For automatic evaluation, we use metrics including BLEU-4~\citep{Papineni2001BleuAM}, CIDEr~\citep{Vedantam2015CIDErCI}, ROUGE-L~\citep{Lin2004ROUGEAP} and METEOR~\citep{Banerjee2005METEORAA} to evaluate the abductive NLG and the explanation generation tasks. We follow common practice in story generation~\citep{Guan2019StoryEG,guan2020knowledge} and use BLEU-1/2 to evaluate the generated endings. We also adopt Distinct-$n$~\citep{Li2016ADO} to measure the diversity of the generated endings.

\subsection{Extracting Sub-Graphs as Knowledge Grounding}\label{sec:subgraph}

To supply knowledge grounding for language generation, we use ConceptNet~\citep{Speer2012RepresentingGR} as the commonsense knowledge base. Each triple $(h, r, t)$ in ConceptNet denotes that the head concept $h$ has a relation $r$ with the tail concept $t$. To condense the knowledge graph $\mathcal{G}=(\mathcal{V}, \mathcal{E})$ we group the original 42 relation types into 17 following \citet{lin-etal-2019-kagnet} and add reversed links $(t, r^{-1}, h)$ to the graph \citep{Lin2018MultiHopKG, Das2018GoFA}.

We extract a sub-graph $G=(V, E)$ from $\mathcal{G}$ which consists of multiple inter-connected paths starting from the source concepts $C_{\bm{x}}$ in the input sequence. To recognize concepts from the input text sequence, we perform fuzzy matching with the lemmatized form of the surface texts using Spacy\footnote{\url{https://spacy.io/}} and filter out stop words. Following \citet{Guan2019StoryEG}, we only consider verbs and nouns as our candidate concepts since we find the extracted graph is much noisier with all the matched concepts.

Specifically, we iterate the following process for $H$ hops: starting from the nodes in the current sub-graph (initialized by $C_{\bm{x}}$) and search for the direct neighbours of each node and preserve top-$B$ nodes with the connected edges to enlarge the sub-graph. For each candidate node, the selection is based on its incoming degree of this node. The incoming degree of a candidate node $v$ is defined as the number of nodes in the current sub-graph that directly connect $v$. Intuitively, we keep those salient concepts that are commonly visited nodes and support information flow on the graph.




\begin{table}[t]
    \centering
    \small
    \begin{tabular}{lccc}
    
    \toprule[1.5pt]
    Tasks & Train & Dev & Test\\
    \midrule[1pt]
        SEG* & 90,000 & 4,081 & 4,081 \\
        $\alpha$NLG & 50,481 & 7,252 & 14,313 \\
        EG* & 25,596 & 1,428 & 2,976 \\
    \bottomrule[1.5pt]
    \end{tabular}
    \caption{Statistics of the datasets used in this paper. *:Examples with multiple references are counted separately.}
    \label{tab:stats}
\end{table}

\subsection{Implementation Details}

\begin{table}[h]
    \centering
    \small
    \begin{tabular}{l ccc}
        \toprule[1.5pt]
        Graph statistics & EG & $\alpha$NLG & SEG
         \\
        \midrule[1pt]
        Avg. \# Concepts & 193.1 & 201.6 & 208.5 \\
        Avg. \# Triples & 1094.3 & 1324.6 & 1148.6 \\
        \bottomrule[1.5pt]
    \end{tabular}
    \caption{Statistics of the extracted subgraphs on the training sets of three datasets, including 
    the average number of concepts and triples for each subgraph.}
    \label{tab:subgraph-stats}
\end{table}

Our model employs the small version of GPT-2 model\footnote{\url{https://github.com/huggingface/transformers}}
with 12 layers, 768-dimensional hidden states, and 12 attention heads 
for contextual modeling and a 2-layer GCN model. We choose the $\texttt{max}(\cdot)$ aggregator for the main results since it yields better performance. During sub-graph extraction, we set the maximum number of hops $H=2$ and preserve top-$B=100$ nodes per hop. We find this setting balances the coverage and noise of the knowledge graph. Detailed statistics of the extracted sub-graphs are presented in Table \ref{tab:subgraph-stats}. To train the model, we use the Adam optimizer~\citep{kingma2014adam} with $\beta_1=0.9, \beta_2=0.999, \varepsilon=1\times10^{-6}$ and linearly decrease learning rate to zero with no warmup.
We search the best hyper-parameters according to BLEU-4 on the development set of each task. At the inference stage, we adopt beam search decoding with a beam size of 3 for our model and all the baselines we produce. We conduct all the experiments using the PyTorch framework~\citep{paszke2017automatic}.

\begin{table*}[t]
    \centering
    \small
    \begin{tabular}{l llll llll}
    \toprule[1.5pt]
    \multirow{2}{*}{Models} & \multicolumn{4}{c}{EG} & \multicolumn{4}{c}{$\alpha$NLG} \\
    \cmidrule(lr){2-5} \cmidrule(lr){6-9}
    {} & BLEU-4 & METEOR & ROUGE-L & CIDEr & BLEU-4 & METEOR & ROUGE-L & CIDEr\\
    \midrule[1pt]
    Seq2Seq & 6.09 & 24.94 & 26.37 & 32.37 & 2.37 & 14.76 & 22.03 & 29.09\\
    COMeT-Txt-GPT2 & N/A & N/A & N/A & N/A & 2.73$^\dagger$ & 18.32$^\dagger$ & 24.39$^\dagger$ & 32.78$^\dagger$ \\
    COMeT-Emb-GPT2 & N/A & N/A & N/A & N/A & 3.66$^\dagger$ & 19.53$^\dagger$ & 24.92$^\dagger$ & 32.67$^\dagger$ \\
    GPT2-FT  & 15.63 & 38.76 & 37.32 & 77.09 & 9.80 & 25.82 & 32.90 & 57.52 \\
    GPT2-OMCS-FT & 15.55 & 38.28 & 37.53 & 75.60 & 9.62 & 25.83 & 32.88 & 57.50 \\
    \midrule[0.5pt]
    GRF & \textbf{17.19} & \textbf{39.15} & \textbf{38.10} & \textbf{81.71}  & \textbf{11.62} & \textbf{27.76} & \textbf{34.62} & \textbf{63.76}\\
    \bottomrule[1.5pt]
    \end{tabular}
    \caption{Automatic evaluation results on the test set of EG and $\alpha$NLG. Entries with N/A mean the baseline is not designated for this task. $\dagger$: we use the generation results from \citet{Bhagavatula2019AbductiveCR}.}
    \label{tab:auto-eval-1}
\end{table*}

\begin{table}[t]
    \centering
    \small
    \begin{tabular}{l c c}
        \toprule[1.5pt]
        Models & BLEU-1/2 & Distinct-2/3 \\
        \midrule[1pt]
        Seq2Seq & 19.1 / 5.5 & 0.181 / 0.360 \\
        IE+GA & 20.8 / 6.4 & 0.140 / 0.280 \\
        WriterForcing &  16.5 / 3.7 & 0.335 / 0.584 \\ 
        GPT2-FT & 25.5 / 10.2 & 0.304 / 0.505 \\ 
        GPT2-OMCS-FT & 25.5 / 10.4 & 0.352 / 0.589 \\
        \midrule[0.5pt]
        GRF &  \textbf{26.1} / \textbf{11.0} & \textbf{0.378} / \textbf{0.622} \\ 
        \bottomrule[1.5pt]
    \end{tabular}
    \caption{Automatic evaluation on the test set of SEG.}
    \label{tab:auto-eval-2}
\end{table}

\begin{table}[h]
    \centering
    \small
    \begin{tabular}{l cc}
        \toprule[1.5pt]
        Models & BLEU-4 & ROUGE-L \\
        \midrule[1pt]
        GRF &  11.62 & 34.62 \\
        w/ $\texttt{mean}(\cdot)$ aggregator & 11.32 & 34.46 \\
        w/o {DMRF}  & 10.67 & 33.75 \\
        w/o {SMGE} & 11.10 & 34.36 \\
        \bottomrule[1.5pt]
    \end{tabular}
    \caption{Ablation study on the test set of $\alpha$NLG. {SMGE} denotes static multi-relational graph encoding (see \S{\ref{sec:static-graph}}) and {DMRF} denotes dynamic multi-hop reasoning
    flow (see \S{\ref{sec:dmrf}}).}
    \label{tab:ablation}
\end{table}

\begin{table*}[t]
    \centering
    \scriptsize
    \begin{tabular}{l llll llll llll}
    \toprule[1.5pt]
        \multirow{3}{*}[-5pt]{Models} & \multicolumn{4}{c}{EG} & 
        \multicolumn{4}{c}{$\alpha$NLG} & \multicolumn{4}{c}{SEG} \\
        \cmidrule(lr){2-5} \cmidrule(lr){6-9} \cmidrule(lr){10-13}
        {} & \multicolumn{2}{c}{Fluency} & \multicolumn{2}{c}{Reasonability} & \multicolumn{2}{c}{Fluency} & \multicolumn{2}{c}{Reasonability} & \multicolumn{2}{c}{Fluency} & \multicolumn{2}{c}{Reasonability} \\
        \cmidrule(lr){2-3} \cmidrule(lr){4-5} \cmidrule(lr){6-7} \cmidrule(lr){8-9} \cmidrule(lr){10-11} \cmidrule(lr){12-13}
        {} & \multicolumn{1}{c}{W} & \multicolumn{1}{c}{L} & \multicolumn{1}{c}{W} & \multicolumn{1}{c}{L} & \multicolumn{1}{c}{W} & \multicolumn{1}{c}{L} & \multicolumn{1}{c}{W} & \multicolumn{1}{c}{L} & \multicolumn{1}{c}{W} & \multicolumn{1}{c}{L} & \multicolumn{1}{c}{W} & \multicolumn{1}{c}{L}\\
    \midrule[1pt]
    vs. IE+GA & N/A & N/A & N/A & N/A & N/A & N/A & N/A & N/A & 0.62** & 0.07 & 0.72** & 0.11 \\
    vs. COMeT-Emb-GPT2 & N/A & N/A & N/A & N/A & 0.31** & 0.14 & 0.55** & 0.25** & N/A & N/A & N/A & N/A \\
    vs. GPT2-FT & 0.24** & 0.09 & 0.54** & 0.21 & 0.15* & 0.10 & 0.56** & 0.20 & 0.21** & 0.12 & 0.45** & 0.19\\
    vs. GPT2-OMCS-FT & 0.18** & 0.09 & 0.58** & 0.18 & 0.12 & 0.09 & 0.50** & 0.20 & 0.17* & 0.11 & 0.40** & 0.15\\
    \bottomrule[1.5pt]
    \end{tabular}
    \caption{Human evaluation results on three datasets. Scores indicate the percentage of \textit{Win} (\textbf{W}) and \textit{Lose} (\textbf{L}) when comparing our model with a baseline in terms of \textit{fluency} and \textit{reasonability}. Scores marked with * mean $\text{p-value}<0.05$ and ** indicates $\text{p-value}<0.01$ in sign test. Entries with N/A mean the baseline is not designated for this task.}
    \label{tab:human}
\end{table*}

\begin{table}[t]
    \centering
    \small
    \begin{tabular}{l c c c}
    \toprule[1.5pt]
    Criteria & \multicolumn{1}{c}{EG} & \multicolumn{1}{c}{$\alpha$NLG} & \multicolumn{1}{c}{SEG} \\
    \midrule[1pt]
    Fluency & 0.615 & 0.543 & 0.315\\
    Reasonability & 0.551 & 0.677 & 0.595 \\
    \bottomrule[1.5pt]
    \end{tabular}
    \caption{Annotator agreement. Scores denotes Fleiss' kappa~\citep{Fleiss1971MeasuringNS} which evaluates the agreement from multiple annotators in terms of \textit{fluency} and \textit{reasonability}.}
    \label{tab:agreement}
\end{table}

\subsection{Compared Baselines}
We produce the following baselines on three generation tasks to compare with our model:

\noindent\textbf{Seq2Seq} is a sequence-to-sequence model based on gated recurrent unit (GRU) with attention mechanism. We also utilize the copying mechanism~\citep{Gu2016IncorporatingCM} for the model to generate out-of-vocabulary words.


\noindent\textbf{GPT2-FT} is a GPT-2 model fine-tuned on the task-specific dataset with its model initialization from \citet{Radford2019LanguageMA}.

\noindent\textbf{GPT2-OMCS-FT} is a commonsense-enhanced GPT-2 model first post-trained on the Open Mind Common Sense (OMCS) corpus\footnote{\url{http://openmind.media.mit.edu}} from which the ConceptNet is constructed. The model is then fine-tuned on the task-specific dataset.

We also compare our model with baseline models designated to each specific task. For story ending generation, we compare to \textbf{IE+GA} which is based on incremental encoding and graph attention~\citep{Guan2019StoryEG} and \textbf{WriterForcing} that forces the attention to focus on important keyphrases and avoid generating generic words. 

For abductive NLG, we compare with two baselines introduced by \citet{Bhagavatula2019AbductiveCR}: \textbf{COMeT-Txt-GPT2} which uses the output texts generated by COMeT as prefix texts to the GPT-2 model while fine-tuning, and \textbf{COMeT-Emb-GPT2} which integrates the embeddings of the outputs generated by COMeT into the GPT-2 model during fine-tuning.


\subsection{Automatic Evaluation}
We present the automatic evaluation results on the test sets of the explanation generation and the abductive NLG tasks in Table \ref{tab:auto-eval-1}. We have the following observations:

\textbf{I.} Our model outperforms all the baselines that utilize pre-trained language models or incorporate external commonsense knowledge in terms of all evaluation metrics indicating that incorporating rich structural information of commonsense knowledge graphs can enhance the overall generation quality.

\textbf{II.} Simply post-training on commonsense knowledge source degrades the performance on these two tasks. This is possibly due to the fact that the triple-level post-trained corpus cannot provide rich semantics for the model to generalize on tasks that emphasize reasoning and explaining.


For story ending generation, we present the evaluation results in Table \ref{tab:auto-eval-2}. Our model outperforms all the baselines in BLEU and distinct metrics. We also observe that post-training on external commonsense data improves the generation diversity of the pre-trained language model, which accords with the findings of \citet{guan2020knowledge}. We suspect that post-training on the commonsense data enables the model to generate concepts related to the story context, which improves the text diversity.

\subsection{Human Evaluation}

To evaluate the fluency and the reasonability of the generated texts under the specific task settings, we conduct pair-wise comparison with \textbf{COMeT-Emb-GPT2} on $\alpha$NLG, \textbf{IE+GA} on SEG, and with two fine-tuned GPT-2 models on all the three tasks. For human evaluation, we randomly sample 100 sentences from the test set for each pair of models and obtain 1,100 sentences from five models. 
We recruit three annotators to make a preference among \textit{win}, \textit{tie} and \textit{lose} given the input context and two outputs generated by our model and a baseline respectively according to two criteria: fluency and reasonability.

For fluency, we require the annotators to focus only on the grammatical correctness and readability of the generated results disregarding the input context. When evaluating reasonability, the annotators are required to assess whether the generated sentence is reasonable under the given context in each task. In SEG and $\alpha$NLG, annotators are asked to focus on evaluating the causal and temporal relevance of the generated results and the contexts. While on EG, annotators are mainly asked to check whether the generated results explain the counterfactual points in the statements properly.


The human evaluation results are presented in Table \ref{tab:human} where our model significantly outperforms compared baselines in terms of both criteria on all the datasets. Specifically, 
incorporating structural commonsense knowledge yields significant improvement in generating reasonable texts given the context. Table \ref{tab:agreement} shows the inter-rater agreement where five out of six annotations show moderate ($0.4\le \kappa <0.6$) or good agreement ($0.6\le \kappa <0.8$). We check the annotation results and find that the GPT-2 baselines also generate story endings with good grammar, which makes it hard for the annotators to reach a high consensus when evaluating the fluency criterion of the story ending generation task ($\kappa = 0.315$).



\subsection{Ablation Study}

We conduct ablation study to verify the effect of different model components. As shown in Table \ref{tab:ablation}, all the components contribute to the final performance. Removing the dynamic reasoning module (w/o DMRF) results in the largest performance drop, thereby indicating that dynamic multi-hop reasoning plays a major role in this task. Ablating the graph representation module (w/o SMGE) also degrades the performance since it encodes the graph structure with relational information that benefits concept selection. We also show the results of our reasoning module with $\texttt{mean}(\cdot)$ aggregator and observe some performance drop comparing with $\texttt{max}(\cdot)$.

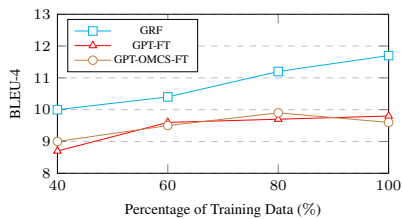
\begin{figure}[t]
\centering
\begin{tikzpicture}[scale=0.8]
\small
\begin{axis}[
    title={},
    width=200pt,
    height=120pt,
    xlabel={Percentage of Training Data ($\%$)},
    ylabel={BLEU-4},
    xmin=40, xmax=100,
    ymin=8, ymax=13,
    xtick={40,60,80,100},
    ytick={8,9,10,11,12,13},
    yticklabel style={font=\scriptsize},
    xticklabel style={font=\scriptsize},
    xlabel style = {font=\scriptsize},
    ylabel style = {font=\scriptsize},
    legend pos=north west,
    legend style={nodes={scale=0.6, transform shape}},
    ymajorgrids=true,
    grid style=dashed
]

\addplot[
    color=cyan,
    mark=square*,
    mark options={fill=white},
    ]
    coordinates {
    (40,10.0)(60,10.4)(80,11.2)(100, 11.7)
    };
    \addlegendentry{GRF}

\addplot[
    color=red,
    mark=triangle*,
    mark options={fill=white},
    ]
    coordinates {
    (40,8.7)(60,9.6)(80,9.7)(100, 9.8)
    };
    \addlegendentry{GPT-FT}

\addplot[
    color=brown,
    mark=*,
    mark options={fill=white},
    ]
    coordinates {
    (40, 9.0) (60,9.5)(80,9.9)(100, 9.6)
    };
    \addlegendentry{GPT-OMCS-FT}

\end{axis}

\end{tikzpicture}
\caption{Performance with different amount of training data on the test set of $\alpha$NLG.}
\label{fig:amount}
\end{figure}

\begin{figure}[t]
\centering
\begin{tikzpicture}[scale=0.8]
\small
\begin{axis}[
    title={},
    width=200pt,
    height=120pt,
    xlabel={$\gamma$},
    ylabel={BLEU-4},
    yticklabel style={font=\scriptsize},
    xticklabel style={font=\scriptsize},
    xlabel style = {font=\scriptsize},
    ylabel style = {font=\scriptsize},
    ymin=10, ymax=12,
    xmin=-0.2, xmax=1,
    ytick={10,10.5,11,11.5,12},
    legend pos=north west,
    ymajorgrids=true,
    grid style=dashed,
    ybar interval=0.7,
]

\addplot
    coordinates {
     (1,11.08) (0.8, 11.15) (0.6, 11.64) (0.4,11.62) (0.2,11.16) (0,10.98) (-0.2,0)
    };

\end{axis}
\end{tikzpicture}
\caption{Effect of $\gamma$ in DMRF. Performance with different value of discount factor $\gamma$ on the test set of $\alpha$NLG.}
\label{fig:gamma}
\end{figure}
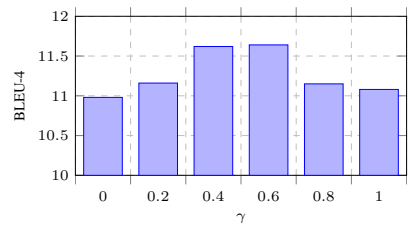

\begin{figure}[t]
    \centering
    \includegraphics[width=0.9\columnwidth]{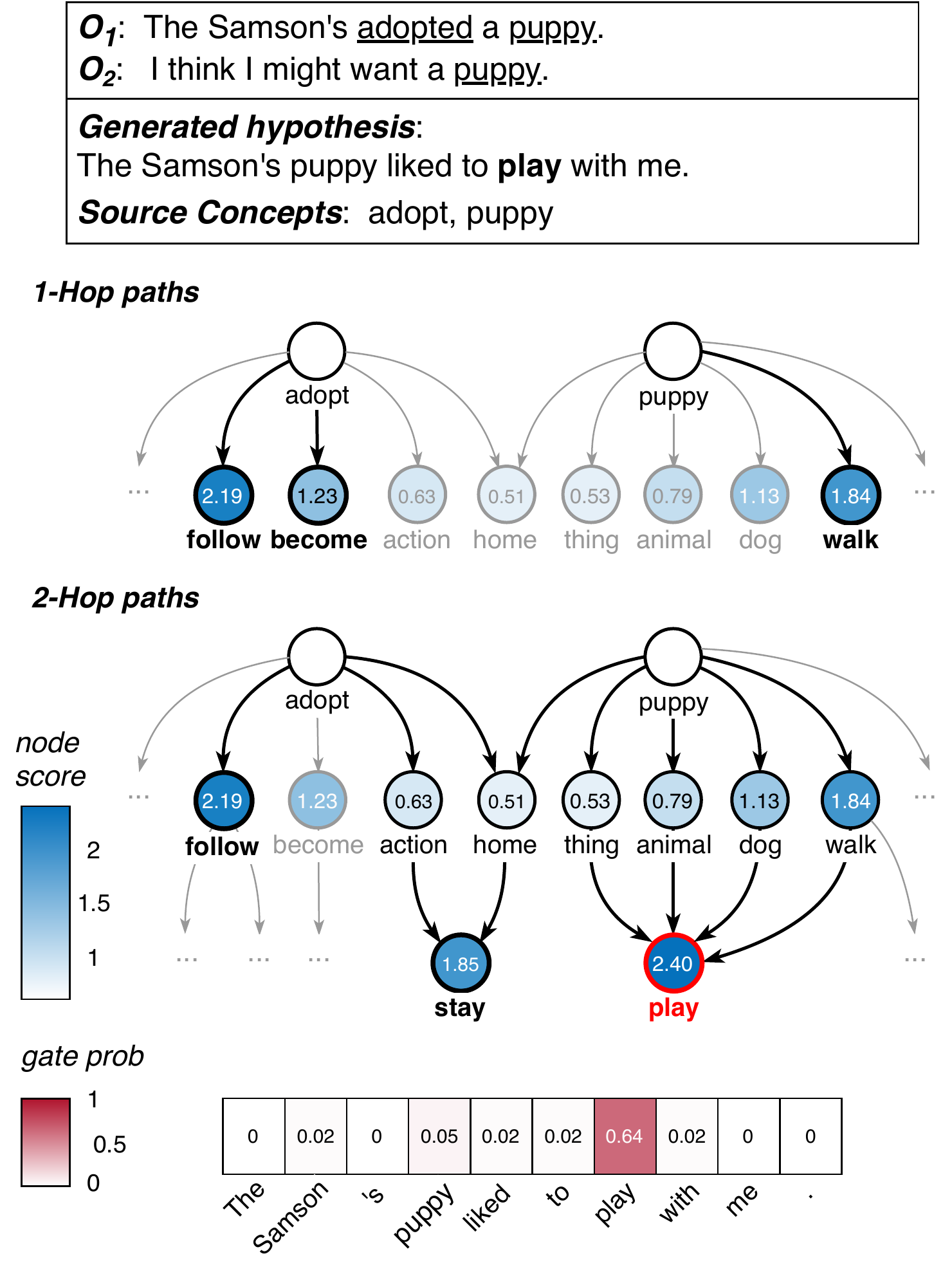}
    \caption{Visualization of a test case with inferred reasoning paths by our model. We highlight top-$3$ concepts with reasoning paths at \textbf{1-Hop} and \textbf{2-Hop} reasoning step respectively.}
    \label{fig:vis}
\end{figure}

\begin{table}[!t]
    \scriptsize
    \centering
    \setlength{\tabcolsep}{0.7mm}{
    \begin{tabular}{l l}
        \toprule[1.5pt]
        \multicolumn{2}{c}{\textbf{Story Ending Generation}} \\
        \midrule[1pt]
        \multirow{3}{*}[-4pt]{\textbf{Story Context}} & \textit{I wanted a simple bike for commuting.} \\
        {} & \textit{So I bought a \textcolor{myblue}{\textbf{cheap}} one on sale.} \\
        {} & \textit{But it didn't fit me properly.} \\
        {} & \textit{And it was uncomfortable to ride.} \\
        \midrule[1pt]
        \textbf{IE+GA} & So I decided to buy a new one. \\
        \textbf{GPT2-FT} & So I decided to buy a new bike. \\
        \textbf{GPT2-OMCS-FT} & So I decided to buy a bike from a bike shop instead. \\
        \textbf{GRF} & So I decided to get a more \textcolor{myorange}{\textbf{expensive}} bike. \\
        \midrule[1pt]
        \multirow{3}{*}[-4pt]{\textbf{Story Context}} & \textit{Ava made shakes for her kids on a \textcolor{myblue}{\textbf{hot}} summer day.} \\
        {} & \textit{She called them in from play, but they dallied.} \\
        {} & \textit{By time they came in, the shakes were almost \textcolor{myblue}{\textbf{melted.}}} \\
        {} & \textit{Ava blended in more \textcolor{myblue}{\textbf{ice}} cubes and refreshed them.} \\
        \midrule[1pt]
        \textbf{IE+GA} & Then she went home and ate them. \\
        \textbf{GPT2-FT} & Ava was proud of her kids for being so good at \\
        {} & cooking. \\
        \textbf{GPT2-OMCS-FT} & She was proud of her kids for being so thoughtful! \\
        \textbf{GRF} & Her kids thanked her profusely for helping them \\
        {} & \textcolor{myorange}{\textbf{cool}} off. \\
        \midrule[1pt]
        \multicolumn{2}{c}{\textbf{Abductive NLG}} \\
        \midrule[1pt]
        \textbf{Observation 1} & \textit{The Smith family went on a \textcolor{myblue}{\textbf{cruise}} for their summer} \\
        {} & \textit{vacation.} \\
        \textbf{Observation 2} & \textit{From then on, the Smiths went to the \textcolor{myblue}{\textbf{beach}} each}\\
        {} & \textit{summer instead.} \\
        \midrule[1pt]
        \textbf{GPT2-FT} & The Smith family had a great time on the beach. \\
        \textbf{GPT2-OMCS-FT} & The Smith family went to the beach. \\
        \textbf{COMeT-Emb-GPT2} & They didn't have a nice vacation. \\
        \textbf{GRF} & The Smith family got \textcolor{myorange}{\textbf{seasick}} on the cruise. \\
        \midrule[1pt]
        \textbf{Observation 1} & \textit{Nancy bought her dog a squeaky stuffed animal.} \\
        \textbf{Observation 2} & \textit{The dog had \textcolor{myblue}{\textbf{ripped}} the toy to shreds.} \\
        \midrule[1pt]
        \textbf{GPT2-FT} & Nancy found a toy that looked like a toy. \\
        \textbf{GPT2-OMCS-FT} & Nancy found a toy that looked like a toy. \\
        \textbf{COMeT-Emb-GPT2} & The squeaky stuffed animal was the first to come in. \\
        \textbf{GRF} & Nancy's dog \textcolor{myorange}{\textbf{scratched}} the stuffed animal. \\
        \midrule[1pt]
        \multicolumn{2}{c}{\textbf{Explanation Generation}} \\
        \midrule[1pt]
        \textbf{Statement} & \textit{\textcolor{myblue}{\textbf{Coke}} is made of \textcolor{myblue}{\textbf{alcohol}}.} \\
        \midrule[1pt]
        \textbf{GPT2-FT} & {Coke is a drink.} \\
        \textbf{GPT2-OMCS-FT} & {Coke is not a liquid.} \\
        \textbf{GRF} & {Coke is made from \textcolor{myorange}{\textbf{corn}}.} \\
        \midrule[1pt]
        \textbf{Statement} & \textit{She \textcolor{myblue}{\textbf{cut}} up a \textcolor{myblue}{\textbf{blanket}}.} \\
        \midrule[1pt]
        \textbf{GPT2-FT} & A blanket is not sharp enough to cut. \\
        \textbf{GPT2-OMCS-FT} & A blanket is too small to be cut. \\
        \textbf{GRF} & Blankets are too \textcolor{myorange}{\textbf{soft}} to be cut. \\
        
        \bottomrule[1.5pt]
    \end{tabular}}
    \caption{Case study on the test set of three datasets. Words in  \textcolor{myblue}{blue} denote source concepts in the input contexts while words in \textcolor{myorange}{orange} are the associated concepts generated by the GRF. }
    \label{tab:case_study}
\end{table}

\subsection{Impact of the Size of Training Data}

To demonstrate the complementary performance gain of utilizing relational paths besides textual modeling, we sample different fractions of training data of $\alpha$NLG for training and evaluate on the original test set. We compare our method with knowledge-agnostic finetuning of the GPT-2 model and the commonsense-enhanced GPT-2 post-trained on OMCS. As shown in Figure \ref{fig:amount}, our model achieves consistent performance gains against the chosen baselines with different amount of training data, which demonstrates the generalization ability of the proposed model with the aid of structural relation knowledge.

\subsection{Effectiveness of Dynamic Multi-Hop Reasoning}

We demonstrate the effectiveness of the multi-hop reasoning module through both quantitative and qualitative analysis. 

We investigate the impact of the hyper-parameter $\gamma$ that controls the information flow in the multi-hop reasoning module. 
As shown in Figure \ref{fig:gamma}, the maximum performance is obtained when $\gamma$ is around $0.4$ and $0.6$. When $\gamma\rightarrow 0$, the multi-hop reasoning module reduces to local scoring of each concept and ignores evidence accumulated on the paths. While $\gamma\rightarrow 1$,
the node score increases monotonically along the paths 
which also hinders the model's ability to select the correct concept. Therefore, we set $\gamma=0.5$ for all the main results of our model.

To qualitatively assess the ability of the dynamic reasoning module, we visualize a test case on $\alpha$NLG with top-ranked concepts and scored reasoning paths. As shown in Figure \ref{fig:vis}, at the first hop the reasoning module starts from the source concepts ``\textit{adopt}'' and ``\textit{puppy}'' and assigns higher scores to neighbouring concepts which are verbs considering the generated context. At the second hop the module utilizes more plausible evidences along 2-hop reasoning paths and selects ``\textit{play}'' ($g_t=0.64$) which is more reasonable given both the observations.



\subsection{Case Study}

We provide some test cases on three datasets in Table \ref{tab:case_study} and observe that: \textbf{I.} Baseline models tend to generate general cases while the GRF is able to generate more specific concepts by exploring the plausible relations between concepts. For example in the first case, the GRF generates ``expensive'' which is the antonym of ``cheap'' under the story context. \textbf{II.} Baseline models sometimes fail to identify the transition of the narrative as shown in the third case where the GRF generates ``seasick'' as a plausible explanation for the transition from ``cruise'' to ``beach''. \textbf{III.} The GRF generates proper attributes of the source concepts in the input context with the aid of external commonsense knowledge as shown in the last two cases of explanation generation.

\section{Conclusion}

We present Generation with Multi-Hop Reasoning Flow that reasons over structured commonsense knowledge during text generation. 
The proposed method leverages both the structural and semantic information of the external knowledge base by performing dynamic multi-hop reasoning on the relational paths. We conduct extensive experiments and empirically show that our method outperforms existing approaches that integrate commonsense knowledge to pre-trained language models on three text generation tasks. We also demonstrate the interpretability of our method with inferred reasoning paths that provide rationale to the generated results.
 
\section*{Acknowledgments}

This work was jointly supported by the NSFC projects (key project with No. 61936010 and regular project with No. 61876096), and the Guoqiang Institute of Tsinghua University with Grant No. 2019GQG1. We thank THUNUS NExT Joint-Lab for the support.

\bibliography{emnlp2020}
\bibliographystyle{acl_natbib}

\end{document}